\documentclass[10pt,twocolumn,letterpaper]{article}

\usepackage{cvpr}              

\usepackage[accsupp]{axessibility} 

\usepackage[dvipsnames]{xcolor}

\definecolor{cvprblue}{rgb}{0.21,0.49,0.74}
\usepackage[pagebackref,breaklinks,colorlinks,citecolor=cvprblue]{hyperref}

\usepackage{algorithm,algorithmic}
\usepackage{color}

\def\paperID{857} 
\def\confName{CVPR}
\def\confYear{2024}

\title{HHMR: Holistic Hand Mesh Recovery by 
\\
Enhancing the Multimodal Controllability of Graph Diffusion Models
}

\author{Mengcheng Li$^{1}$, Hongwen Zhang$^{2}$, Yuxiang Zhang$^{1}$, Ruizhi Shao$^{1}$, Tao Yu$^{3}$, Yebin Liu$^{1}$\footnotemark[1]. \\
$^1$Department of Automation, Tsinghua University \\
$^2$School of Artificial Intelligence, Beijing Normal University. \\
$^3$Beijing National Research Center for Information Science and Technology, Tsinghua University
}

\begin{document}

\twocolumn[{
\maketitle
 \begin{figure}[H]
 \vspace{-20pt}
 \hsize=\textwidth
    \centering
    \includegraphics[width=2\linewidth]{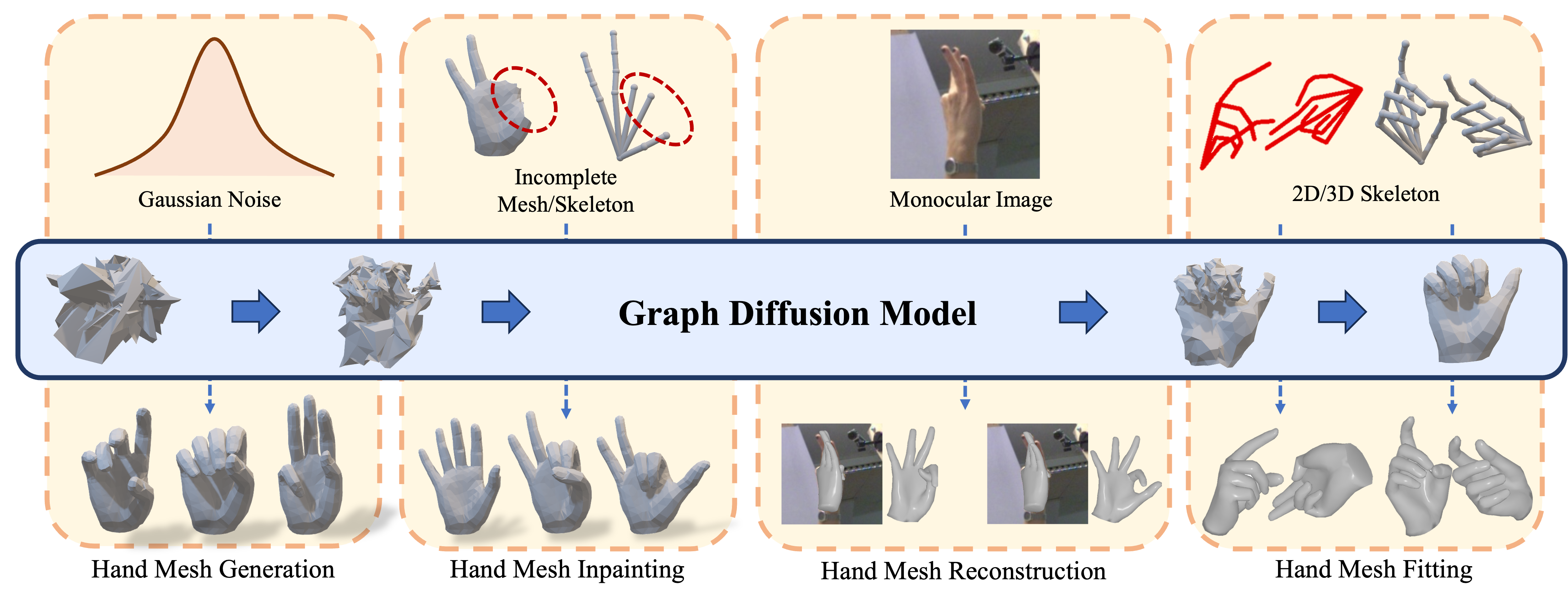}
    \caption{We introduce \textbf{HHMR}, a graph diffusion-based generation framework that are compatible with various human hand mesh recovery tasks.
    }
    \label{fig:teaser}
\end{figure}
}]

\renewcommand{\thefootnote}{\fnsymbol{footnote}} 
\footnotetext[1]{Corresponding Author.} 

\begin{abstract}

Recent years have witnessed a trend of the deep integration of the generation and reconstruction paradigms. 
In this paper, we extend the ability of controllable generative models for a more comprehensive hand mesh recovery task: direct hand mesh generation, inpainting, reconstruction, and fitting in a single framework, which we name as \textbf{H}olistic \textbf{H}and \textbf{M}esh \textbf{R}ecovery (HHMR).
Our key observation is that different kinds of hand mesh recovery tasks can be achieved by a single generative model with strong multimodal controllability, and in such a framework, realizing different tasks only requires giving different signals as conditions.
To achieve this goal, we propose an all-in-one diffusion framework based on graph convolution and attention mechanisms for holistic hand mesh recovery. 
In order to achieve strong control generation capability while ensuring the decoupling of multimodal control signals, we map different modalities to a shared feature space and apply cross-scale random masking in both modality and feature levels.
In this way, the correlation between different modalities can be fully exploited during the learning of hand priors.
Furthermore, we propose Condition-aligned Gradient Guidance to enhance the alignment of the generated model with the control signals, which significantly improves the accuracy of the hand mesh reconstruction and fitting.
Experiments show that our novel framework can realize multiple hand mesh recovery tasks simultaneously and outperform the existing methods in different tasks, which provides more possibilities for subsequent downstream applications including gesture recognition, pose generation, mesh editing, and so on.

\end{abstract}    
\section{Introduction}
\label{sec:intro}

Hand mesh recovery aims to recover hand pose and mesh from images, which has achieved a wide range of applications in ARVR, human-computer interaction, robotics, and etc,. 
In the past decades, with the development of deep learning techniques, the paradigm of hand mesh recovery has undergone a shift from keypoints-based mesh\&model fitting \cite{taylor2017articulated,hoyet2012sleight,zhao2012combining,wang2020rgb2hands,gao20223d} to learning-based regression. 
Regression-based hand mesh recovery methods are based on the encoder-decoder architecture which first encodes various kinds of inputs, usually images or keypoints, and then decodes to either 3D poses \cite{spurr2020biomechanical}, pose and shape parameters \cite{chen2021model,zhou2020monocular,boukhayma20193d,Zhang_2019_ICCV} of a parametric hand model \cite{MANO:SIGGRAPHASIA:2017} or the vertex coordinates of a 3D hand mesh directly \cite{Ge_2019_CVPR,chen2021cameraspace,tang2021towards,chen2022mobrecon,kulon2020weakly,lin2021mesh,lin2021metro}. 
However, the one-way regression strategy does not establish or fully utilize the underlying distributions of natural hand pose and mesh, which significantly restricts the achievement of more advanced hand mesh recovery tasks, such as mesh\&pose inpainting or generation. 

Despite the classical hand mesh recovery task, another spectrum of research focuses on formulating the priors (or distributions) of natural hand pose or mesh by low-dimensional manifolds \cite{tiwari2022posendf}. 
Acquiring such a prior can not only facilitate the hand mesh recovery but also enable the generation of hands that ideally follow the biomechanical constraints. 
Classical hand priors either rely on hand-crafted 3D hand models with restricted joint angles \cite{spurr2020biomechanical} and degrees of freedom (DoFs) \cite{Yang_2021_ICCV_CPF}.
Romero \etal \cite{MANO:SIGGRAPHASIA:2017} introduced an explicit parametric hand model by principal component analysis (PCA) on a dataset with natural hand pose and shape parameters. 
However, the limited representative power of explicit constraints or PCA significantly limits the ability to represent complex postures not to say generate diverse hand meshes.  


Benefiting from the rapid development of generative learning techniques,  generation-based implicit hand models become a hot topic. 
Some works \cite{zuo2023twohandvae,li2021task} have utilized the Variational Autoencoder (VAE) to map the plausible hand parameters to a Gaussian space. For the first time, it becomes possible to generate hand meshes through sampling in feasible distributions. 
Implicit hand models can also be used for hand mesh recovery. By using two VAEs to encode the image domain and hand pose domain independently, CrossVAE or Generative Adversarial Networks (GAN) are incorporated to align the two domains and enable the generation of hand parameters that align with the input image (\cite{wan2017crossing,spurr2018cross,yang2019aligning,yang2019disentangling}). 
However, existing implicit hand models are mostly built on the parameter space or 3D skeleton space, which makes it hard to generate meshes or utilize geometric constraints directly. Moreover, additional inversion or fitting processes are necessary for hand mesh recovery based on existing implicit hand models, which is time-consuming and sophisticated. 

An inevitable trend for related research is the unification of hand mesh recovery and the generative hand models. 
This will enhance the mutual benefits between existing reconstruction and generation paradigms and finally enable direct hand mesh generation, inpainting, reconstruction, and fitting simultaneously in an End2End manner. 
We call such a comprehensive hand mesh recovery task as \textbf{H}olistic \textbf{H}and \textbf{M}esh \textbf{R}ecovery, named HHMR, and propose an All-In-One diffusion framework based on GCN and attention for holistic hand mesh recovery without any additional finetuning or inversion steps. 
Our key insight is that conditional generation based on diffusion models, through carefully multimodal designed and conditioned training, is a powerful methodology for achieving holistic hand mesh recovery. 
Given different modalities of conditions as input, the diffusion model can produce plausible hand mesh and pose results thus naturally supports:  
i) hand mesh reconstruction when given image as condition, 
ii) hand mesh inpainting when given incomplete mesh or 2D\&3D skeleton as condition, 
iii) hand mesh generation when given random noise as input (with no condition), 
and iv) hand mesh fitting when given a 2D skeleton as the condition.  
To fulfill the accurate multimodal controllability of the diffusion models, we map different conditions to a shared feature space and apply a random mask strategy at both modality and feature levels to enhance the correlation learning between different modalities.
Moreover, we propose a condition-aligned gradient guidance strategy during diffusion learning to further improve the alignment with the conditions. 

We evaluate the proposed holistic hand mesh recovery framework on various downstream tasks. Empirical results demonstrate that i) on hand mesh generation task, our approach generates more diverse poses. ii) On hand mesh inpainting task, our method can recover multiple plausible hand meshes from incomplete inputs. iii) Our method achieved comparable results with SOTAs on single-hypothesis reconstruction and outperforms SOTAs on multi-hypothesis reconstruction tasks. iv) With condition-aligned gradient guidance, our method achieves performance with higher precision in 2D fitting tasks.


The contributions can be summarized as: 
\begin{itemize}
\item We propose an All-In-One framework based on a graph diffusion model for holistic hand mesh recovery. The hand prior learned in the graph diffusion model can be readily applied to different downstream tasks without any additional finetuning or inversion steps.
\item We map different modality conditions to a shared feature space and apply a random mask strategy at both modality and feature levels to enhance the correlation learning between different modalities.
\item We propose a condition-aligned gradient guidance strategy during inference to further improve the alignment with the conditions. 
\item Extensive experiments and comparisons validate the effectiveness of our framework and demonstrate various downstream applications. 
\end{itemize}
\section{Related Work}

\subsection{Hand Mesh Recovery}

In the past few decades, significant progress has been made in human hand recovery. Early research \cite{taylor2017articulated,hoyet2012sleight,zhao2012combining,wang2020rgb2hands,gao20223d,lightcap2021} utilized optimization methods to fit the hand pose from 2D skeletons detection.

With the development of neural networks, some learning-based hand mesh reconstruction methods have been proposed. One approach is utilizing the popular parameter-based MANO \cite{MANO:SIGGRAPHASIA:2017} model and regressing the pose and shape parameters of it. Due to the differentiability of the MANO model, it becomes feasible to estimate model parameters end-to-end from a single-view image input (\cite{chen2021model,zhou2020monocular,boukhayma20193d,Zhang_2019_ICCV,pymafx2023,zhang2023proxycap}). However, parameter estimation is a highly nonlinear task that lacks the correlation with the 3D space. Another approach is directly regressing hand mesh, which is typically achieved by building a mapping network between 2D image input and 3D hand mesh output. Graph convolution network (GCN) is one of the widely employed mapping networks (\cite{Ge_2019_CVPR,chen2021cameraspace,tang2021towards,chen2022mobrecon,kulon2020weakly,Li2022intaghand}). Additionally, there are methods that utilize one-dimensional lixel heatmaps \cite{moon2020i2l} to represent the 3D coordinates of vertices, or employ UV map \cite{chen2021i2uv} to establish connections between 2D and 3D. Lin \etal \cite{lin2021metro,lin2021mesh} used transformer-based network to regress vertices. 

However, these hand reconstruction methods only provide a single plausible estimation, whereas in reality, the occluded parts could assume various poses. Thus, we propose a probability diffusion-based model that generates multiple plausible hand meshes from one input RGB image.


\subsection{Hand Prior and Generation}

The hand prior model is aimed at learning a distribution of plausible hand poses. Previous prior models can be broadly categorized into two types: one type learns the unconditional distribution $p_{data}(x)$ of the human hand, and the other is the task-specific prior that learning the conditional distribution $p_{data}(x|c)$ under given conditions (such as RGB image or 2D skeleton).

For the unconditional prior distribution, one straightforward approach is to manually constrain the reasonable range of hand poses based on the biological structure of the human hand. Yang \etal \cite{Yang_2021_ICCV_CPF} have constructed the hand pose prior by manually defining the degrees of freedom for each joint and the range of rotation angles. Spurr \etal \cite{spurr2020biomechanical} proposed a set of losses that constrain hand pose to lie within the range of biomechanically feasible 3D hand configurations. The other line of works is learning-based which involves learning feasible distribution from a large dataset of poses. Javier \etal \cite{MANO:SIGGRAPHASIA:2017} applied principal component analysis (PCA) on the pose and shape parameters of a hand dataset. In both human hand domain \cite{zuo2023twohandvae,li2021task} and body domain \cite{pavlakos2019SMPLX}, researches have been conducted to map the pose distribution to a standard Gaussian distribution by an VAE network. Tiwari \etal \cite{tiwari2022posendf} described the prior of the human body by learning the neural distance from the sampled pose to the low-dimensional manifold of reasonable pose distributions. These unconditional priors often require optimization when applied in downstream applications, which can be time-consuming.

The conditional prior model often learns the potential pose distribution under specific task constraints, such as predicting plausible pose from RGB images or 2D skeletal. A common approach is to construct VAEs in different domains (\eg depth image, RGB image or 3D skeleton) and then build hand priors through latent space alignment(\cite{wan2017crossing,spurr2018cross,yang2019aligning,yang2019disentangling}). Recently, Ci \etal \cite{ci2023gfpose} predicted the gradient of log-likelihood of the human body pose distribution through a score matching based model. However, these implicit conditional prior models are mostly built on the 3D skeletons and rely on specific conditions, which restricts their application.

\subsection{Diffusion based generative model}

The diffusion model \cite{sohl2015deep} is a generative model based on stochastic diffusion processes, which uses a Markov chain to gradually convert one distribution into another as modeled in Thermodynamics. In practice, it pre-defines a forward process that gradually adds noise to a dataset distribution until it converges to a standard Gaussian noise distribution. Subsequently, a neural network is trained to learn the reverse process and gradually denoise a random noise into a reasonable data sample. Recently, some image generation algorithms \cite{DDPM,DDIM,stable_diffusion} that utilize diffusion models have achieved significant breakthroughs. Diffusion models have also demonstrated great success in the fields of 3D object generation \cite{poole2022dreamfusion,ruiz2023dreambooth,lin2023magic3d}, human pose estimation \cite{feng2023diffpose,gong2023diffpose,holmquist2023diffpose}, and human motion generation \cite{MDM,priorMDM}. We build a diffusion-based hand generation model with a classifier-free strategy, which can run with or without conditional. Hence, it is capable of handling various downstream tasks.

\section{Preliminary: Diffusion Model} \label{sec:diff}

The denoising diffusion model is a probabilistic generative model that consists of a forward process and a reverse process. We denote the middle distribution for each step in the process as $p(x_t), t=0,1,...,T$, where the initial $p(x_0)$ is the dataset distribution we want to generate, and the final $p(x_T)\sim\mathcal{N}(\mathbf{0}, \mathbf{I})$ is a standard Gaussian noise distribution.









The forward process converts $p(x_0)$ to $p(x_T)$ by gradually adding small Gaussian noises with predefined means and variances in a Markov Chain:
\begin{equation}
    p(x_t|x_{t-1}) = \mathcal{N}(\sqrt{1-\beta_t}x_{t-1}, \beta_t \mathbf{I})
\end{equation}
where $\{\beta_t\}$ is a predefined set of small constants.

The reverse process is a generation process that converts the noise sampled from $p(x_T)$ to a reasonable data sample in $p(x_0)$. The reverse process is also a Gaussian distribution \cite{feller2015retracted}. Moreover, the reverse diffusion process can be written as:

\begin{equation}
    p_{\mathbf{\theta}}(x_{t-1}|x_t, \mathbf{c}) = \mathcal{N}(\mu_{t}^{\theta}(x_t, t, \mathbf{c}), \Sigma_t \mathbf{I})
\end{equation}

where $\mathbf{c}$ represents the conditions for generation and $\mu_{t}^{\theta}$, $\Sigma_t$ are means and variances of the reverse Gaussian process. Usually, $\Sigma_t$ is determined by $\{\beta_t\}$, while $\mu_{t}$ is untrackable. Thus, a neural network $\mu_{t}^{\theta}(x_t, t, \mathbf{c})$ is applied to predict the means $\mu_{t}$.

Furthermore, for a more effective learning of $\mu_{t}^{\theta}$, it is typically reparameterized to learn i) the noise $\epsilon$ added from $x_0$ to $x_t$ or ii) $x_0$ instead. Image generation tasks \cite{DDPM,DDIM,ho2022classifierfree,dhariwal2021diffusion_classifer_guidance} usually employ the first setting, whereas we follow the human motion generation methods \cite{priorMDM,MDM} to predict $x_0$ directly:
\begin{equation}
    \Bar{x}_0 = f_\mathbf{\theta}(x_t, \mathbf{c}, t), \quad \mu_{t}^{\theta}(x_t, t, \mathbf{c})=g_t(\Bar{x}_0, x_t, t),
\end{equation}
where $f$ is a neural network, $\mathbf{\theta}$ represents the training parameters, $g_t$ is a determined reparameterized function. More details can be referred to \cite{DDPM}.

In summary, the training algorithm for the diffusion model involves four parts: i) sample data $x_0$ from dataset $p(x_0)$, ii) run the forward process to add noise on $x_0$ and yield $x_t$, iii) run the neural network $f_\mathbf{\theta}$ to give a prediction of $\Bar{x_0}$, iv) back propagate the prediction loss $\|\Bar{x}_0 - x_0\|$.

\section{Method}

\begin{figure*}
    \centering
    \includegraphics[width=0.95\linewidth]{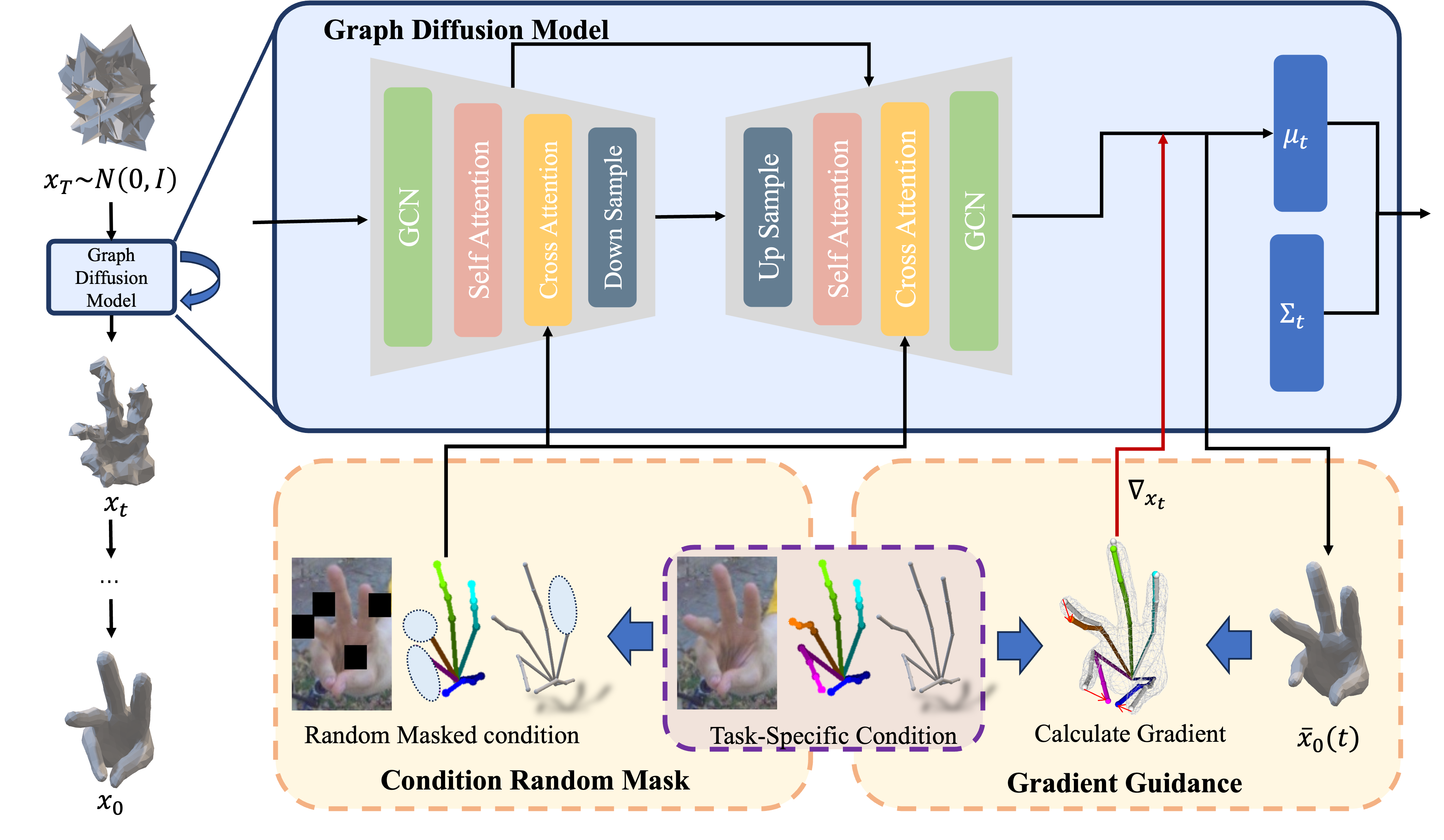}
    \caption{The pipeline of our graph diffusion model. With task-specific conditions, our model progressively removes noise from randomly Gaussian noise and directly reconstructs the complete hand meshes. Additionally, we introduce a gradient-based guidance to improve the alignment between the generated results and observations.
    }
    \label{fig:pipeline}
    \vspace{0pt}
\end{figure*}

\subsection{Problem Statement}\label{subsec:cgdm}

Our goal is to develop a diffusion model that learns a prior knowledge of the human hand from the given task-specific conditions $f_{\theta}(x_t, \mathbf{c}, t)$. Considering the highly non-linear mapping from the parameter space of the hand template to 3D spatial space, we directly process on the 3D meshes instead. We drew inspiration from the network architecture of image generation method \cite{stable_diffusion}, but employed 3D graph convolutions network (GCN) \cite{defferrard2016gcn} instead of 2D image convolutions to encode the local information of the human hand mesh. The details of the network architecture will be presented in the Sec.\ref{subsec:hand_model}.

In order to accommodate different downstream tasks, our approach supports a variety of conditions $\mathbf{c}$ as input. Specifically, we set $\mathbf{c}$ to a RGB image $\mathbf{c}_{image}$ to handle monocular hand mesh reconstruction task; set $\mathbf{c}$ to 2D skeleton $\mathbf{c}_{skel2d}$ for 2D hand mesh fitting task; set $\mathbf{c}$ to incomplete 3D skeleton $\mathbf{c}_{skel3d}$ for hand mesh inpainting task. We can also set $c=\emptyset \empty$ to achieve unconditional generation, and apply it to the task of hand postures generation and hand mesh completion.

Due to the different input conditions, our network can be trained using different types of datasets. For the dataset containing paired human hand images and annotated hand meshes, we can train our network under all the aforementioned conditions: $\mathbf{c}=\mathbf{c}_{image} | \mathbf{c}_{skel2d} | \mathbf{c}_{skel3d}$. For the dataset containing only human hand poses, we employ skeletal information $\mathbf{c}=\mathbf{c}_{skel2d} | \mathbf{c}_{skel3d}$ as the condition to train our generation model.

\subsection{Graph Diffusion Network}\label{subsec:hand_model}


\paragraph{Network architecture.}
As shown in Fig.~\ref{fig:pipeline}, the main structure of our diffusion model is a U-shaped network, consisting of a downsampling encoder and an upsampling decoder with skip connections. Each block of both encoder and decoder are formed by 4 layers: \ie, GCN layer, self-attention layer, cross-attention layer, and optional upsampling or downsampling layer.

Similar to the convolution in 2D image networks, we use GCN~\cite{defferrard2016gcn} to aggregate information from each vertex's neighbors to encode local information of hand mesh, and expand the receptive field through stacking multiple layers of GCN. Then a self-attention layer is utilized to facilitate long-range global information propagation on the 3D hand mesh, enabling our prior network to construct global relationships across the surface vertices. We also apply a cross-attention layer to inject conditional information $\textbf{c}$ into our generative model.

\paragraph{Condition Mapping.}
In order to encode different conditions into the same space that are compatible with the input of the cross-attention layer, we designed distinct networks to map different conditions to a shared feature space:
\begin{itemize}
\item For 3D skeleton and 2D skeleton condition, we simply use MLP to encode $\mathbf{c}_{skel2d}$ and $\mathbf{c}_{skel3d}$. 
\item For RGB image condition $\textbf{c}_{image}$, we first use a CNN-based network to encode the latent image feature map and then crop it into several image patches. After that, a Vision Transformer (ViT) \cite{dosovitskiy2020vit} based model is applied to further process image information. 
We also add a global token $\textbf{c}_{global}$ while running ViT to encode the global information as the original ViT does. The global token is further stacked with the output image patch features $\textbf{c}_{patch}$ of ViT to form the image condition $\textbf{c}_{image} \in \mathcal{R}^{(P+1)\times D}$, where $P$ is the number of image patches. 
\end{itemize}

\paragraph{Conditions with Random Mask.}
\label{subsec:mask}
In our method, we stack all three types of condition $\mathbf{c} = \mathbf{c}_{image}\oplus\mathbf{c}_{skel2d}\oplus\mathbf{c}_{skel3d}\in \mathcal{R}^{(P+1+1+1)\times D}$ as the input of cross-attention layers.
To enhance the network's generalization and versatility, we design a two-level masking strategy. 
On the multimodal level, in order to decouple the relationship between conditions, we independently set each condition $\mathbf{c}_{image} | \mathbf{c}_{skel2d} | \mathbf{c}_{skel3d}$ to an empty set $\emptyset$ with probability $p_m$ during training. Furthermore, we also globally mask all conditions with probability $p_{all}$ to train our model's unconditional generation capability.
On the feature level, We design different masking strategies for each condition individually. For image condition $\textbf{c}_{image}$, we encode it into several image patches and randomly mask each patch with probability $p_{image}$, which enables our method to handle complex occlusion input. For skeleton condition $\textbf{c}_{skel2d}$ and $\textbf{c}_{skel3d}$, we add Gaussian noise on each joint and randomly mask each finger with probability $p_{skel}$, which facilitates the generation from incomplete hand skeletons.

Note that when applying our model to a specific task, the irrelevant conditions can be directly set to $\emptyset$. For example, in the monocular reconstruction task, the condition is $\mathbf{c} = \mathbf{c}_{image}\oplus\emptyset\oplus\emptyset$, which we write as $\mathbf{c} = \mathbf{c}_{image}$ for short.

\subsection{Condition-aligned Gradient Guidance}\label{subsec:g_guidance}
In the 2D image generation task, a classifier guidance~\cite{dhariwal2021diffusion_classifer_guidance} approach is proposed for the conditional diffusion model. It trains an additional classifier network to predict the probability of input condition from $x_t$. Then, by modifying the means of the reverse process with a gradient of the log-likelihood predicted from the classifier network, the diffusion model can generate images that are more consistent with the input conditions.

Inspired by previous work~\cite{dhariwal2021diffusion_classifer_guidance}, we propose a Condition-aligned Gradient Guidance to encourage that the generated hand is consistent with the input condition. Specifically, in our implementation, we add a gradient guidance bias to the means of each Gaussian distribution of the reverse process:
\begin{equation}
\label{equ:cg}
    \bar{\mu}_t = \mu_t - s \Sigma_t \triangledown_{x_t}{\|Pf_\mathbf{\theta}(x_t, \mathbf{c}, t) - Px_0\|}
\end{equation}
where $s$ is a scale factor and $P$ is a task-specific operator. Noted that $f_\mathbf{\theta}(x_t, \mathbf{c}, t)$ is the prediction of the GT mesh $x_0$, this method can be seen as a form of neural fitting. The operator $P$ determines the specific fitting target. For example, by setting $P$ as a joint regression matrix $\mathcal{J}$, the gradient guidance works as a supervise term to constrain the 3D joints of generated hand meshes to be consistent with the 3D skeleton condition.

In addition to serving as a supervisory, this gradient-based guidance approach can also be viewed as a geometric control during the generation process. For example, in the hand mesh inpainting task, given an incomplete hand mesh, we can apply the gradient guidance on the given part and leave the missing part uncontrolled. In this situation, the operator $P$ is a per-vertex binary mask $M_V$ that indicates the missing part of hand meshes. We can also use partial skeletons to control the generation of the human hand by setting $P$ as $M_J\mathcal{J}$, where $M_J$ is a per-joint binary mask.

Note that in contrast to the original classifier guidance method, our gradient guidance approach does not need any extra classifier network. Hence, our approach requires no additional training, which is less time-consuming and plug-to-use. 

\subsection{Loss Functions}
\label{subsec:loss}

For training our graph diffusion model, we utilize (1) diffusion data loss, (2) vertex loss \& joint loss and (3) mesh smooth loss. 

\noindent\textbf{Data Loss.} We use L1 loss to supervise the output of our diffusion model as described in Sec.\ref{sec:diff}:
\begin{equation}
    L_{data} = \|x_0 - \bar{x}_0\|
\end{equation}

\noindent\textbf{Vertex Loss \& Joint Loss.} We pretrained a joint regression matrix $\mathcal{J}$ to regress joints from output hand mesh and apply L1 loss to supervise 3D coordinates of vertices and joints:

\begin{equation}
\begin{aligned}
    L_{V} & = \sum\|V- V^{GT}\|_1 \\
    L_{J} & = \sum\|\mathcal{J}V- \mathcal{J}V^{GT}\|_1 
\end{aligned}
\end{equation}

\noindent\textbf{Mesh Smooth Loss.} To ensure the geometric smoothness of the predicted vertices, two different smooth losses are applied. First, we regularize the normal consistency between the predicted and the ground truth mesh:
\begin{equation}
\begin{aligned}\textbf{}
    L_{n} & = \sum\|e \cdot n^{GT}\|_1,
\end{aligned}
\end{equation}
where $e$ is the edges of the generated hand mesh and $n^{GT}$ is the faces normal vector calculated from the ground truth mesh. 
Also, we minimize the L1 distance of each edge length between the generated mesh and the ground truth mesh:
\begin{equation}
\begin{aligned}\textbf{}
    L_{e} & = \sum\|e - e^{GT}\|_1. 
\end{aligned}
\end{equation}

\section{Experiments}

\begin{figure*}
    \centering
    \includegraphics[width=0.95\linewidth]{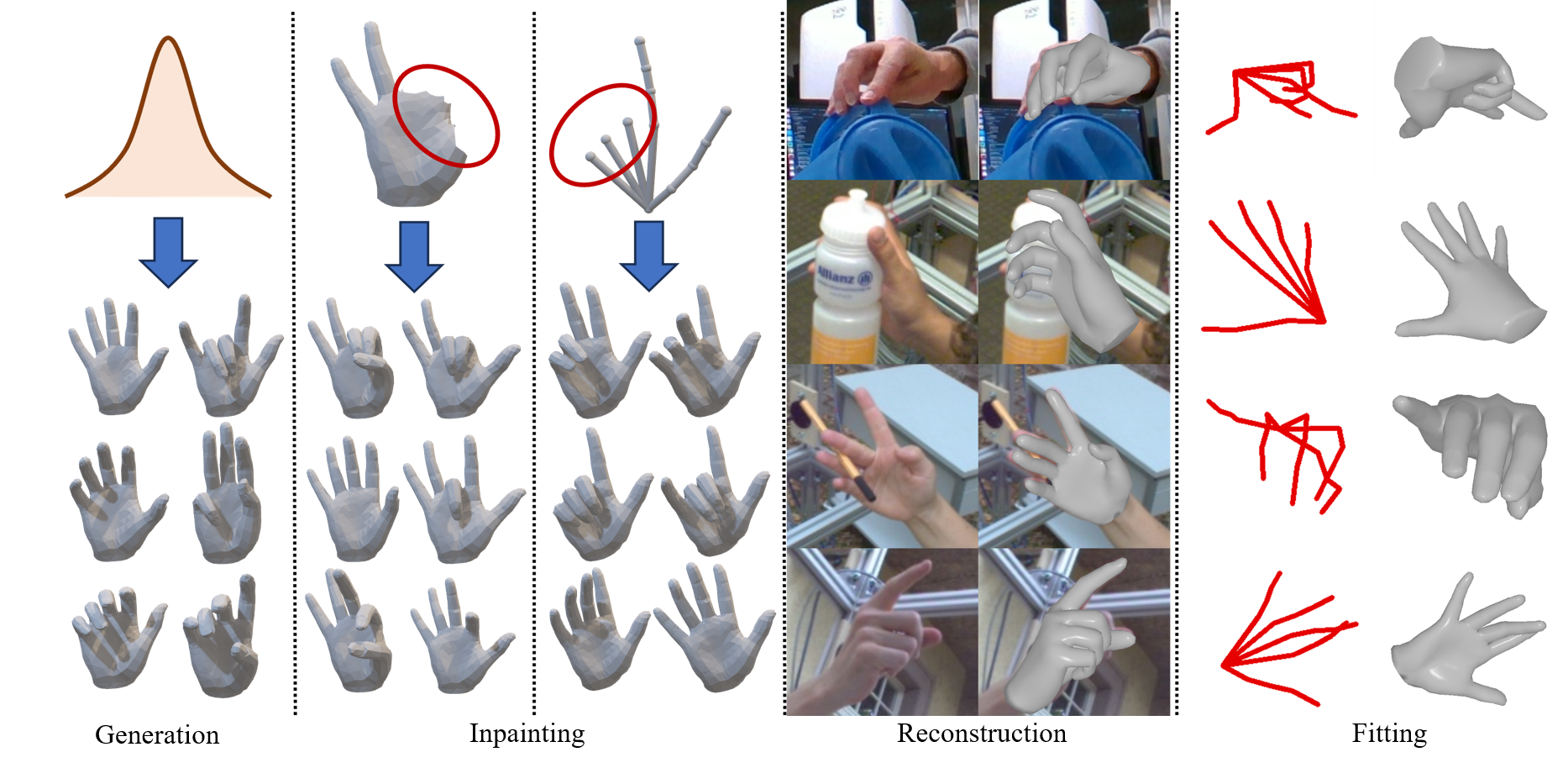}
    \caption{Qualitative results of our method for different downstream tasks. From left to right are i) hand mesh generation results from random Gaussian noise, ii) hand mesh inpainting from incomplete hand mesh or skeleton, iii) hand mesh reconstruction from monocular RGB image, and iv) hand mesh fitting from 2D skeletons.
    }
    \label{fig:main_result}
    \vspace{-0pt}
\end{figure*}

\subsection{Experimental Setting}
\label{subsec:exp_set}

\noindent\textbf{Implementation Details.} Our network is implemented by PyTorch. For RGB image condition, we use ResNet50 \cite{he2016resnet} with initial weights pretrained on ImageNet \cite{deng2009imagenet} as the backbone for the mapping network and then evenly crop the image feature into $8\times8$ patches. For 2D/3D skeleton conditions, we use a three-layer MLP with dropout \cite{hinton2012dropout} and GELU \cite{hendrycks2016gelu} activation. For the diffusion model, we set the denoising steps to 1000 during training, and utilized the DDIM \cite{DDIM} algorithm to speed up inference.

\noindent\textbf{Training Datasets.} We simultaneously leveraged hand pose datasets with only hand pose annotations and reconstruction datasets with both RGB images and hand pose annotations. For pose datasets, we use Two-hand 500K \cite{zuo2023twohandvae} and InterHand2.6M \cite{Moon_2020_ECCV_InterHand2.6M} datasets. Two-hand 500K \cite{zuo2023twohandvae} is a two-hand pose dataset that utlized hand instances sampled from single-hand datasets \cite{2020FreiHAND,Moon_2020_ECCV_InterHand2.6M,zimmermann2017learning,zhang2017stb}. InterHand2.6M \cite{Moon_2020_ECCV_InterHand2.6M} contains both single-hand and two hands motion sequence data in 30FPS. we split the pose data of two hands into two individual single-hand data. For reconstruction datasets, we use FreiHand \cite{2020FreiHAND}, HO3D\_V3 \cite{hampali2020honnotate,hampali2022keypointtransformer}, CompHand \cite{chen2021cameraspace,chen2022mobrecon}, both of them contain hand images and hand mesh annotations.

\noindent\textbf{Training Details.} We train our model with a mixture of the above datasets. We train our model using the AdamW optimizer \cite{adamw} on a single NVIDIA RTX 4090 GPU with a mini-batch size 64. We trained for 1M mini-batches in total. The learning rate is setting to $5e^{-4}$ and decrease to $1e^{-5}$ with cosine annealing.

\subsection{Hand Mesh Generation}
\label{subsec:exp_gen}
Our model can work without any condition by setting $\mathbf{c}=\emptyset$. In this setting, our approach involves sampling a set of noisy point clouds from a Gaussian distribution and denoising them into a complete human hand mesh as shown in Fig.\ref{fig:main_result}. To quantitatively evaluate our generative model, we follow Tiwari \etal \cite{tiwari2022posendf} to use Average Pairwise Distance (APD) to quantify the diversity of generated meshes and calculate the percentage of self-intersecting faces (SI) to evaluate the realism.
We randomly generate $n=500$ hand meshes with our model and randomly sample $n=500$ meshes within the PCA space of hand parameters from MANO \cite{MANO:SIGGRAPHASIA:2017}. 
As shown in Tab.\ref{tab:gen}, our results exhibit higher APD and lower SI compared to the PCA sampling, indicating that our method is capable of producing more diverse and realistic hand meshes.

\begin{table}
\centering
\begin{tabular}{lcc}
\hline
     & APD($mm$)$\uparrow$ & SI($\%$)$\downarrow$ \\ 
\hline
PCA Generation        & 15.00 & 0.39  \\   
Ours Generation & \textbf{17.30} & \textbf{0.05} \\
\hline
PCA Inpainting & 12.28 & 0.06 \\
Ours Inpainting & \textbf{13.59} & \textbf{0.01} \\
\hline
\end{tabular}
\caption{Quantitatively comparison of hand mesh generation and inpainting.}
\label{tab:gen}
\vspace{-15pt}
\end{table}

\subsection{Hand Mesh Inpainting}
\label{subsec:exp_inpaint}
Given an incomplete hand mesh or hand 3D skeleton, our model can inpaint the whole plausible hand mesh according to the prior knowledge learned from a large hand pose dataset. The key idea of mesh inpainting is using the gradient guidance to keep the given part of the hand unchanged while allowing the diffusion model to give various generations for the missing part. As illustrated in the Fig.\ref{fig:main_result}, where the thumb, index, and middle fingers are all missing from the hand mesh, our model can provide various potential completion results. Also, given a 3D hand skeleton that missing the middle, ring, and little fingers, our model can generate diverse whole-hand meshes. Please refer to the Supplementary for the details of the mesh inpainting algorithm. 
We also conducted quantitative evaluation. As shown in Tab.\ref{tab:gen}, our method achieved a higher APD and a lower SI compared to randomly sampling by PCA.

        

\subsubsection{Hand Mesh Reconstruction}


\begin{figure}
    \centering
    \includegraphics[width=0.95\linewidth]{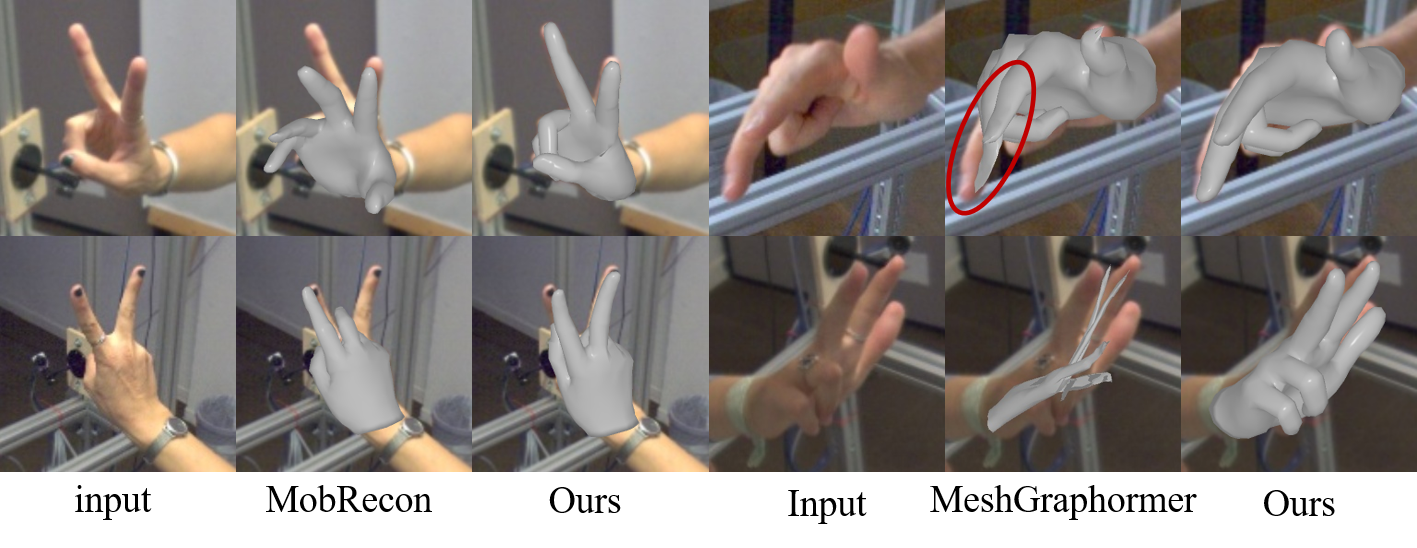}
    \caption{Qualitative results comparison with MobRecon \cite{chen2022mobrecon} and MeshGraphormer \cite{lin2021mesh}.
    }
    \label{fig:comp}
    \vspace{-15pt}
\end{figure}

\noindent\textbf{Single-Hypothesis.}
Given an RGB image as condition $\mathbf{c}_{image}$, our model can work as a monocular hand mesh reconstruction network. We first quantitatively evaluate our method on Freihand \cite{2020FreiHAND} dataset. We use DDIM \cite{DDIM} algorithm to run $T=10$ denoising steps and only sample one hypothesis. Note that we follow MeshGraphormer \cite{lin2021mesh} to use the test time augmentation during evaluation. We report the Mean Per Joint Position Error (MPJPE) and the Mean Per Vertex Position Error (MPVPE) after rigid alignment in Tab.\ref{tab:sh}, it can be seen that our method can achieve comparable results with state-of-the-art (SOTA) methods for single-hypothesis reconstruction (n=1). We then demonstrate qualitative comparison results on the Freihand dataset in Fig.\ref{fig:comp}, our method performs better on side viewpoint situations and occlusions situations. We believe that this is due to the fact that our model is a prior model, which can better recover hand meshes from incomplete observations.
We also conducted ablation study on the usage of GCN module. We modify our model by replacing all GCNs with self-attention layers (named as ``Only SA") and report results in Tab.\ref{tab:as}, where our method is clearly better.

\begin{table}
\centering
\begin{tabular}{lcc}
\hline
     & MPJPE & MPVPE \\ 
\hline
Ours (\textbf{w} GCN) & \textbf{5.99} & \textbf{5.97} \\
Only SA (\textbf{w/o} GCN)       & 6.39 & 6.35 \\
\hline
\end{tabular}
\caption{Ablation study of using GCN module.}
\vspace{-25pt}
\label{tab:as}
\end{table}

\begin{table}
\centering
\begin{tabular}{l|cc}
\hline
     & MPJPE & MPVPE \\ 
\hline
FreiHAND \cite{2020FreiHAND}      & 11.0 & 10.9 \\
YotubeHand \cite{kulon2020weakly} & 8.4  & 8.6 \\
I2L-MeshNet  \cite{moon2020i2l}   & 7.4  & 7.6 \\
HIU-DMTL \cite{zhang2021hand}     & 7.1  & 7.3 \\
CMR \cite{chen2021cameraspace}    & 16.9 & 7.0 \\
I2UV-HandNet \cite{chen2021i2uv}  & 6.7  & 6.9 \\
METRO \cite{lin2021metro}         & 6.7  & 6.8 \\
Tang \etal \cite{tang2021towards} & 6.7  & 6.7 \\
MobRecon \cite{chen2022mobrecon}  & 6.1  & 6.2 \\
MeshGraphormer \cite{lin2021mesh} & 5.9  & 6.0 \\
\hline
Ours (n=1)                & 6.0 & 6.0 \\
Ours (n=8)                & 5.9 & 5.9 \\
Ours (n=16)               & \textbf{5.8} & \textbf{5.8} \\
\hline
\end{tabular}
\caption{Quantitative results on the Freihand dataset. Our method applied $T=10$ steps denoising by DDIM \cite{DDIM} and $n$ represents the number of samples.}
\label{tab:sh}
\vspace{-15pt}
\end{table}

\begin{figure}
    \centering
    \includegraphics[width=0.95\linewidth]{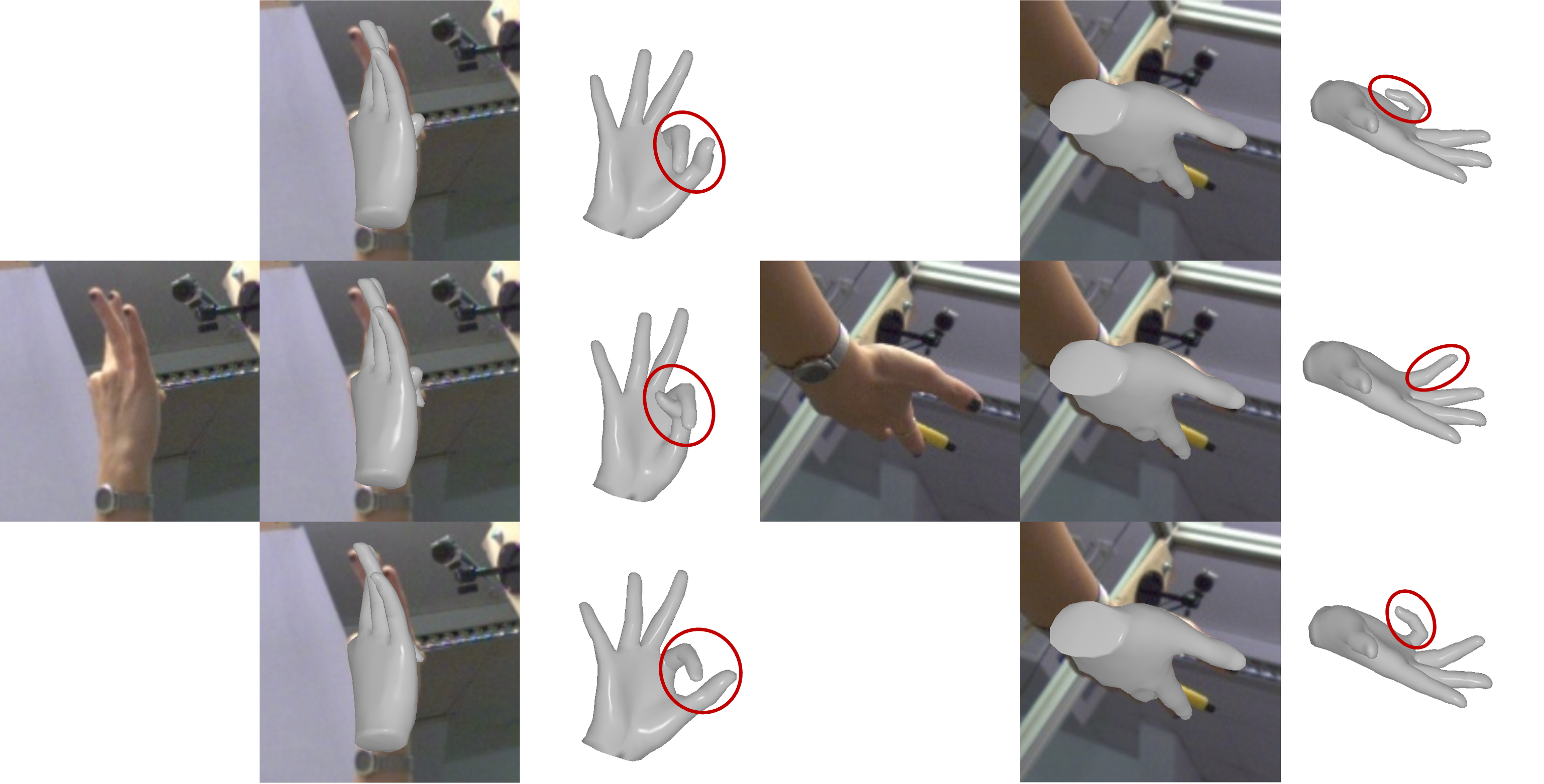}
    \caption{Results for multi-hypothesis reconstruction from monocular image.
    }
    \label{fig:prob}
    \vspace{-0pt}
\end{figure}

\noindent\textbf{Multi-Hypothesis.} By denoising different noises sampled from Gaussian distribution, our method is capable of providing multiple mesh reconstruction hypotheses from a single image. We first qualitatively demonstrate the multi-hypothesis results in Fig.\ref{fig:prob}, showing that our method can maintain alignment with the input image in visible areas while offering multiple potential guesses for the occluded regions. We then quantitatively analyze the effect of the hypothesis count in Tab.\ref{tab:mh}. We also evaluate the effect of inference steps. Following human body multi-hypothesis methods \cite{ci2023gfpose, kolotouros2021prohmr}, we report the minimum MPJPE and MPVPE. As shown in Tab.\ref{tab:mh} and Tab.\ref{tab:sh}, when only 8 samples are drawn, our method already outperforms the SOTAs \cite{chen2022mobrecon, lin2021mesh}. 
It is also indicated in Tab.\ref{tab:mh} that satisfactory results can be achieved with just $T=10$ steps of denoising.

\begin{table}
\centering
\begin{tabular}{ll|cc}
\hline
     &       &  MPJPE  &  MPVPE  \\ 
\hline
T=10 & n=8   & 5.91 & 5.90 \\
     & n=16  & 5.76 & 5.76 \\
     & n=32  & 5.63 & 5.64 \\
\hline
T=25 & n=8   & 5.79 & 5.79 \\
     & n=16  & 5.58 & 5.59 \\
     & n=32  & 5.41 & 5.43 \\
\hline
T=50 & n=8   & 5.73 & 5.73 \\
     & n=16  & 5.51 & 5.52 \\
     & n=32  & \textbf{5.33} & \textbf{5.35} \\
\hline
\end{tabular}
\caption{Effect of denoising steps and sampling quantity. $T$ represents the number of denoising steps by DDIM. $n$ represents the number of samples.}
\label{tab:mh}
\vspace{-0pt}
\end{table}

\subsubsection{Hand Mesh Fitting}

Our method can also take a 2D skeleton as a condition $\mathbf{c}_{skel2d}$, and can also utilize the gradient guidance approach to further enhance alignment. We set the operator $P$ in Equ.\ref{equ:cg} as $P=M\Pi\mathcal{J}$, where $\mathcal{J}$ is a pre-trained skeleton regression matrix, $\Pi$ is the projection operator, $M$ is a per-joint mask that masks out 2D joints with low detection confidence. We employ the ground truth 2D skeleton as supervision for quantitative evaluation under various inputs.

We firstly compare with the traditional 2D skeleton fitting algorithm by utilizing the hand PCA prior \cite{MANO:SIGGRAPHASIA:2017}. We utilize 2D skeleton $\mathbf{c}_{skel2D}$ as input conditional and also apply the gradient guidance $\triangledown_{skel2D}$. As demonstrated by the second and third rows in Tab.\ref{tab:fit2d}, our method achieved higher fitting accuracy. This is because our method is a 3D prior model that is capable of learning the relationship from 2D skeleton to 3D hand mesh. 

\begin{table}
\centering
\begin{tabular}{l|cc}
\hline
           &  MPVPE  &  MPJPE  \\ 
\hline
Fitting 2D skeleton &  7.00  &  7.27  \\
$\mathbf{c}_{skel2D} + \triangledown_{skel2D}$           &  6.28  &  6.26  \\ 
\hline
$\mathbf{c}_{image}$                                                     &  6.54  &  6.57  \\ 
$\mathbf{c}_{image} + \mathbf{c}_{skel2D}$                               &  5.32  &  5.29  \\ 
$\mathbf{c}_{image} + \mathbf{c}_{skel2D} + \triangledown_{skel2D}$      &  \textbf{5.05}  &  \textbf{4.94}  \\
\hline
\end{tabular}
\caption{Quantitative results for hand mesh fitting, $\mathbf{c}_{image}$ represents input RGB image, $\mathbf{c}_{skel2D}$ represents input 2D skeleton, $\triangledown_{skel2D}$ represents applying the gradient guidance on 2D skeleton}
\label{tab:fit2d}
\vspace{-0pt}
\end{table}

We also introduce the 2D skeleton information $\mathbf{c}_{skel2D}$ into the hand mesh reconstruction task with image input $\textbf{c}_{image}$ to further improve the accuracy by 2D fitting. Quantitative results in Tab.\ref{tab:fit2d} demonstrate that our fitting method can further improve the alignment with the input image compared to the reconstruction results. please refer to the Supplementary Materials for more qualitative results.


\section{Discussion}

\noindent\textbf{Conclusion.} 
We introduce a graph diffusion based generative network that is compatible with various downstream hand mesh recovery tasks, including hand mesh generation, hand mesh inpainting, hand mesh reconstruction and hand mesh fitting. Our network can take different task-specific conditions as input, and directly denoise a 3D hand mesh from randomly sampled noisy point clouds. We also designed a conditional mask training strategy to address missing and noisy conditional inputs, and further employed a gradient guidance method for enhancing the consistency between the generation output and the input conditions.

\noindent\textbf{Limitation.}
Although our method uses a mask training strategy to work with noisy and incomplete conditional inputs, it may still fail for extremely missing or excessively noisy input conditions. Additionally, while our method only requires $T=10$ denoising steps to achieve decent results, increasing the denoising steps for more precise generation is time-consuming. 

\noindent\textbf{Acknowledgement}: This paper is supported by National Key R\&D Program of China (2022YFF0902200), the NSFC project No.62125107, the Beijing Municipal Science \& Technology Z231100005923030.
{
    \small
    \bibliographystyle{ieeenat_fullname}
    \bibliography{ref,ref_intag,ref_mvhand}
}


\end{document}